\documentclass[11pt,a4paper]{article}
\usepackage[hyperref]{eacl2021}
\usepackage{times}
\usepackage{latexsym}

\usepackage{microtype}

\aclfinalcopy 

\usepackage{times}
\usepackage{latexsym}
\usepackage{soul}

\usepackage{url}
\usepackage{xcolor}
\usepackage{hyperref}
\usepackage{multirow}
\usepackage{graphicx}
\usepackage{booktabs}
\usepackage{caption}
\usepackage{subcaption}
\usepackage[ruled,vlined]{algorithm2e}
\usepackage{colortbl}

\newcommand{\mn}[1]{\textcolor{black}{#1}}

\newcommand{\fix}[1]{\textcolor{black}{#1}}

\title{Self-Learning for Zero Shot Neural Machine Translation}

\makeatletter
\def\@fnsymbol#1{\ensuremath{\ifcase#1\or \dagger\or \ddagger\or
   \mathsection\or \mathparagraph\or \|\or **\or \dagger\dagger
   \or \ddagger\ddagger \else\@ctrerr\fi}}
\makeatother
\author{ Surafel M. Lakew\thanks{$^{\dagger}$Work conducted when the author was at FBK.},
Matteo Negri, Marco Turchi \\
  $^{\dagger}$University of Trento, Fondazione Bruno Kessler, Trento, Italy \\
  \texttt{surname@fbk.eu}
}

\date{}

\begin{document}
\maketitle
\begin{abstract}
Neural  Machine Translation (NMT) approaches employing monolingual data are showing steady improvements in resource-rich conditions.
However, evaluations using real-world low-resource languages still result in unsatisfactory performance. This work proposes a novel \textit{zero-shot} NMT modeling approach that learns without the now-standard assumption of a \textit{pivot} language sharing parallel data with the zero-shot \textit{source} and \textit{target} languages.~Our approach is based on three stages:~\textit{initialization} from any pre-trained NMT model observing at least the target language, \textit{augmentation} of source sides leveraging target monolingual data, and \textit{learning} to optimize the initial model to the zero-shot pair, where the latter two constitute a self-learning cycle. Empirical findings involving four diverse (in terms of a language family, script and relatedness) zero-shot pairs show the effectiveness of our approach 
with up to +$5.93$ BLEU improvement against a supervised bilingual baseline. Compared to unsupervised NMT, consistent improvements are observed even in a domain-mismatch setting, attesting to the usability of our method.
\end{abstract}

\section{Introduction}\label{sec:intro}
\begin{figure}[!t]
    \centering 
    \begin{subfigure}{0.135\textwidth}
      \includegraphics[width=\linewidth]{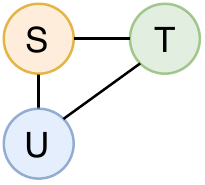}
      \caption{Multilingual} 
      \label{fig:mnmt}
    \end{subfigure} \hspace{2em}
    \begin{subfigure}{0.135\textwidth}
      \includegraphics[width=\linewidth]{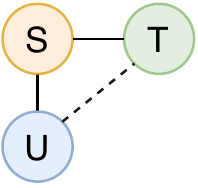}
      \caption{Zero-Shot} 
      \label{fig:zst}
    \end{subfigure}

    \medskip
    \begin{subfigure}{0.136\textwidth}
      \includegraphics[width=\linewidth]{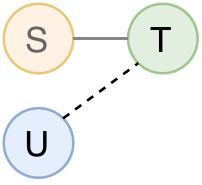}
      \caption{Zero-Shot} 
      \label{fig:ourZNMT}
    \end{subfigure} \hspace{2em}
    \begin{subfigure}{0.136\textwidth}
      \includegraphics[width=\linewidth]{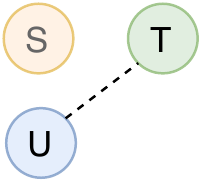}
      \caption{Unsupervised} 
      \label{fig:unmt}
    \end{subfigure}
\caption{Proposed (c) and existing NMT approaches using parallel (\textit{solid} line) and monolingual (\textit{broken line}) data.}
\label{fig:zNMT-unmt-settings}
\end{figure}
Since the introduction of NMT~\cite{sutskever2014sequence,bahdanau2014neural}, model learning using unlabeled (\textit{monolingual}) data is increasingly gaining ground. Undoubtedly, the main motivating factor to explore beyond \emph{supervised} learning is the lack of enough (\textit{parallel}) examples, a performance bottleneck regardless of the underlying architecture~\cite{koehn2017six}.
A fairly successful approach using monolingual data is the \emph{semi-supervised} learning with back-translation~\cite{sennrich2015improvingMono}, particularly if the initial supervised model is good enough for augmenting quality pseudo-bitext~\cite{poncelas2018investigating,ott2018scaling,caswell2019tagged}. Moreover, back-translation showed to be a core element of new monolingual based approaches.
These include zero-shot NMT~\cite{lakew2017improving,gu2019improved,currey2019zero}, which relies on a multilingual model~\cite{johnson2016google,ha2016toward} (Fig.~\ref{fig:zst}) and unsupervised NMT, which initializes from pre-trained embeddings~\cite{lample2017unsupervised,aretxe2018unsup} or cross-lingual language model~\cite{lample2019cross} (Fig.~\ref{fig:unmt}).
At least two observations can be made on the approaches that leverage monolingual data: 
{\it i}) they require high-quality and comparable monolingual examples, and 
{\it ii}) they show poor performance on real-world zero-resource language pairs ({\sc zrp}s)\footnote{{\sc zrp}: a language pair with only monolingual data available, alternatively called Zero-Shot Pair ({\sc zsp}).}~\cite{neubig18:rapid,guzman2019two}.

To overcome these problems, in this work we propose a zero-shot modeling approach (Fig.~\ref{fig:ourZNMT}) to translate from an unseen \textit{source}  language $U$ to a \textit{target} language $T$ that has only been observed by a model pre-trained on ($S,T$)
parallel data ($S$ being a different source language). In literature, zero-shot NMT has been investigated under the assumption that ($U,P$) and ($P,T$) parallel data involving a \textit{pivot} language $P$ are available for pre-training (Fig.~\ref{fig:zst})~\cite{johnson2016google,ha2016toward}. However, most of the +$7,000$  currently spoken languages do not exhibit any parallel data with a common $P$ language. \emph{This calls for new techniques to achieve zero-shot NMT by reducing the requirements of the pivoting-based method}. 

To this aim, our approach follows a self-learning cycle, by first translating in the \textit{primal} zero-shot direction $U \rightarrow T$ with a model pre-trained on (\textit{S, T}) parallel data that has never seen $U$ during training. Then, the generated translations are used as a pseudo-bitext to learn a model for the \textit{dual} $T\rightarrow U$ translation direction. This inference-learning cycle is iterated, alternating the dual and primal zero-shot directions until convergence of the $U\leftrightarrow T$ zero-shot model.

Through experiments on data covering eight language directions,
we demonstrate the effectiveness of our approach in the {\sc zrp} scenario.
In particular, we report significant improvements compared to both a \textit{supervised} model trained in 
low-resource conditions (up to $+5.93$ BLEU) and an \textit{unsupervised} one exploiting large multilingual corpora (up to $+5.23$ BLEU).
\\

\noindent
Our contributions can be summarized as follows:
\begin{itemize}
    \item We propose a new variant of zero-shot NMT, which reduces the requirements of previous pivoting-based methods. Our approach enables incorporating an unseen zero-resource language $U$, with no need of pre-training on parallel data involving $U$.

    \item We empirically evaluate our approach on diverse language directions and  in a real-world zero-resource scenario, a testing condition disregarded in previous literature.

    \item We provide a rigorous comparison against unsupervised neural machine translation, by testing our models in an in-domain, out-of-domain, and source to target domain mismatch scenarios.
\end{itemize}

\section{Background and Motivation}
\label{sec:background}
For an $S$ and $T$ language pair, a standard NMT model is learned by mapping ($s,t$) example pairs with an encoder-decoder network~\cite{kalchbrenner2013recurrent,sutskever2014sequence} such as Recurrent~\cite{bahdanau2014neural,cho2014learningGRU}, Convolutional~\cite{gehring2017convolutional}, and Transformer~\cite{vaswani2017attention}. Despite the varied architectural choices, the objective of NMT is to minimize the loss,
\begin{equation}\label{eq:nmt-mle}
    L_{\hat{\theta}} (s,t) = \sum_{i=1}^{|t|+1} log p(t_i|s,\hat{\theta}) 
\end{equation}
$\hat{\theta}$ is the parameterization of the network, $s$ is the source sentence, and 
$t$ is the predicted sentence. \fix{Reserved tokens $\langle bos \rangle$ at $i=0$ and $\langle eos \rangle$ at $i=|t|+1$ defines the beginning and end of $t$.}

\subsection{Training Paradigms}\label{subsec:train-paradigms}
\noindent
{\bf Semi-Supervised} learning leverages monolingual data and has been used to improve supervised phrase-based models~\cite{bertoldi2009domain,bojar2011improving}. In NMT the procedure is commonly called $-$ {\it back-translation}~\cite{sennrich2015improvingMono}; to enhance $S\rightarrow T$ direction additional pseudo-bitext is utilized by augmenting the $S$ side from $T$ monolingual segments with a reverse $T\rightarrow S$ model. Back-translation became a core module in approaches that leverage monolingual data, such as dual-learning~\cite{dual-learningMT,sestorain2018zero}, zero-shot~\cite{firat2016zero,lakew2017improving,gu2019improved,currey2019zero,zhang2020improving}, and unsupervised~\cite{lample2017unsupervised,aretxe2018unsup} translation.
\\

\noindent
{\bf Unsupervised} learning considers only monolingual data of {\bf {\it S}} and {\bf {\it T}} languages. Initialization from pre-trained embeddings~\cite{aretxe2018unsup,lample2017unsupervised} or cross-lingual language model~\cite{lample2019cross}, denoising auto-encoder and iterative back-translation are commonly employed learning objectives. 
Despite being a rapidly growing research \fix{area}, findings show failures in an unsupervised NMT when using real-world {\sc zrp}~\cite{neubig18:rapid}, distant languages~\cite{guzman2019two} in a domain-mismatched scenario~\cite{kim2020and,artetxe2020call}. Given the similarity in leveraging monolingual data, we directly compare our zero-shot NMT with unsupervised NMT.
\\

\noindent
{\bf Multilingual} modeling extends Eq.~\ref{eq:nmt-mle} objective to multiple language pairs. Although early work dedicates network components per language~\cite{dong2015multi,luong2015multi,firat2016multi}, the most effective way utilizes ``{\it language-id}'' to share a single encoder-decoder model across multiple language pairs~\cite{johnson2016google,ha2016toward}. For $L$ languages, a model learns to maximize the likelihood over all the available language pairs (max of $N=L(L-1)$) parallel data. For each language pair $S,T \in N$ and $S \neq T$, Eq.~\ref{eq:nmt-mle} can be written as,
\begin{equation}\label{eq:mnmt-mle}
    L_{\hat{\theta}}^T (s^S,t^T) = \sum_{i=1}^{|t|+1} log p( t_i^T|s^S,\hat{\theta} )
\end{equation}
Where the \textit{language-id} (i.e. $\langle 2T \rangle$) is explicitly inserted at \fix{$i$=1 of the source ($s^T$)}. 
Most importantly, multilingual modeling enables translation between language pairs without an actual training data ($S,T \notin N$), exploiting an implicit transfer-learning from pairs 
with training data $-$ also known as \textit{Zero-Shot Translation} (\textsc{zst})~\cite{johnson2016google}.

\fix{
Transfer Learning across NMT models (i.e, \textit{parent-to-child})~\cite{zoph16:tf}, have been shown to work effectively with \mn{a} shared vocabulary across related~\cite{nguyen2017transfer} and even distant~\cite{kocmi2018trivial} languages, by pre-training multilingual  models~\cite{neubig18:rapid}, by updating parent vocabularies with child~\cite{lakew18:tl-dv}, and for a \textsc{zst} with a pivot language~\cite{kim2019pivot}.  In this work, we leverage \mn{for the first time} pre-trained models for zero-shot translation without the pivot language assumption.    
}

\subsection{Zero-Shot Translation}
From a broad perspective, {\sc zst} research is moving in three directions, 
(\textit{i}) improving translation quality by employing {\sc zst} specific objectives~\cite{chen2017teacher,lu2018neural,blackwood2018multilingual,al2019consistency,arivazhagan2019missing,pham2019improving,ji2019cross,siddhant2020leveraging}, (\textit{ii}) training favorable large scale multilingual models for the {\sc zst}  languages with lexically and linguistically similar languages~\cite{aharoni2019massively,arivazhagan2019massively}, and
(\textit{iii}) \mn{incrementally learning}
better model for the {\sc zst} directions 
with self-learning objectives \cite{lakew2017improving,gu2019improved,zhang2020improving}. 
The common way of employing \textit{self-supervised learning} in {\sc zst} modeling is iterative back-translation that generates the source from the monolingual target to construct a new parallel sentence pair. In terms of performance, while (\textit{i}) and (\textit{ii}) fall behind,   (\textit{iii}) either approaches or even outperform\mn{s} the two-step \textit{pivot translation} approach ($S\rightarrow P\rightarrow T$).

Despite these progress\fix{es}, current approaches make identical \mn{assumptions, namely: \textit{i) the}} reliance on a multilingual model\mn{,} and \mn{\textit{ii)}} observing both the $U$ and $T$ zero-shot languages paired with the $P$ language(s). 
However, in a real-world setting these conditions are rarely satisfied, hindering the application of zero-shot NMT to the vast majority of {\sc zrp}. Moreover, conditioning {\sc zst} on $P$ language(s) creates a performance ceiling that depends on the amount, domain, and quality of parallel data available for the $S-P$ and $P-T$ pairs.

To our knowledge, a zero-shot NMT modeling between an \textit{unseen}-$U$ and $T$ zero-shot pair, and without the {\it P} language(s) criterion has not yet been explored, motivating this work.

\section{Zero-Shot Neural Machine Translation}\label{sec:znmt} \mn{We}
propose a new zero-shot NMT ({\sc zNMT}) approach that expands the current definition of zero-shot to the extreme scenario,  where we avoid the established assumption of observing the zero-shot ($U$, $T$) languages paired with the pivot ($P$) language(s). Instead, \mn{we consider}
only the availability of monolingual data for ($U$, $T$) and a pre-trained NMT observing only the $T$ language $-$ a scenario applicable to most zero-resource languages.

To this end, with the goal of learning a zero-shot model covering the primal ($U\rightarrow T$) and the dual ($T\rightarrow U$) directions, our \textsc{zNMT} approach consists of three stages: model initialization, incremental data augmentation, and model learning. 
The latter two steps can be iterated over time creating a \textit{self-learning cycle} between the primal and dual zero-shot directions.

\begin{algorithm}[!t]
\small
\SetAlgoLined
 \nl 
 Input\;
    $U_m$, $T_m$ $\leftarrow$  monolingual data of the {\sc zsp}\;
    {\sc TM} $\leftarrow$  pre-trained translation model\;
    $R$ $\leftarrow$ maximum self-learning cyle\;
 \nl {\sc zNMT}$^0$, $\leftarrow$ {\sc TM}\;
 \nl $D_{Pr}$ = $\emptyset$; $D_{Du}$ = $\emptyset$\;
 \nl $r \leftarrow 1$ \;  
 \nl\For{$r$ \KwTo $R$}{

    \nl $T^{*} \leftarrow Primal\_Infer$ ({\sc zNMT}$^0$, $U_m$)
    
    \nl $D_{Du}$ $\leftarrow$ ($T^{*}$, $U_m$)
    
    \nl {\sc zNMT}$^{1}$ $\leftarrow$ Train ({\sc zNMT}$^0$, $D_{Du} \cup D_{Pr}$)
    
    \nl $D_{Pr}$ = $\emptyset$ 
    
    \nl $U^{*} \leftarrow Dual\_Infer$ ({\sc zNMT}$^{1}$, $T_m$)
    
    \nl $D_{Pr}$ $\leftarrow$ ($U^{*}$, $T_m$)
    
    \nl {\sc zNMT}$^{2}$ $\leftarrow$ Train ({\sc zNMT}$^{1}$, $D_{Pr} \cup D_{Du}$)
    
    \nl {\sc zNMT}$^0$ $\leftarrow$ {\sc zNMT}$^{2}$ 
     
    \nl $D_{Du}$ = $\emptyset$ 
    }
   \nl {\bf return} {\sc zNMT}$^0$
 \caption{Zero-Shot NMT Modeling}\label{algo:zNMT}
\end{algorithm}

\subsection{Model Initialization}
Different from conventional {\sc zst} approaches, \textsc{zNMT} can be either initialized from a bilingual or a multilingual pre-trained system (Algorithm~\ref{algo:zNMT},~\textit{line}~2). 
The only assumption we consider is the availability of the zero-shot $T$ side at pre-training time. Hence, our initialization introduces relaxation to model pre-training, a direct consequence of removing the $P$ language premise. Considering that $U$ has never been observed, we analyze different pre-training strategies to build a robust zero-shot system.

\subsection{Self-Learning Cycle} 
In our scenario, we have only access to monolingual data for both the $U$ and $T$ languages, and the pre-trained model did not leverage $U - T$ parallel data during training. In this setting, after the initialization, the very first task is a zero-shot inference for $U\rightarrow T$ (\textit{line} 6), \mn{to which we refer} as the \textit{primal} {\sc zst} direction. The goal of this step is to acquire pseudo-bitext\mn{s} to enhance the translation capability of the NMT system for the $T\rightarrow U$ direction.

\mn{To this aim,} when the primal inference is concluded, a learning step is performed by reverting the generated pseudo-bitext data ($T\rightarrow U$).  \textsc{zNMT} optimizes the same objective function (Eq.~\ref{eq:nmt-mle}) as in the pre-training or Eq.~\ref{eq:mnmt-mle} if zero-shot training is performed together with other supervised languages pairs (\textit{co-learning}) as in~\cite{johnson2016google}. The \mn{resulting} model is then used to perform the $T \rightarrow U$ inference (\textit{line} 10), \mn{to which we refer}
as the \textit{dual} direction. The data generated by the dual process is then paired with the data produced by the primal \mn{one} and used in a new leaning step. Assuming that at each inference step our algorithm generates better quality bi-text\mn{s}, we replace the dual  or primal data ($D_{Du}$ and $D_{Pr}$) produced at the previous round with the ones generate\mn{d} in the current round. For instance, during the training at line 8, we use the $D_{Du}$  generated at line 7, while we keep the $D_{Pr}$ generated  at line 11 of the previous round.

This sequential approach alternates the primal and dual inferences (\textit{lines} 6, 10) with a learning phase in between (\textit{lines} 8, 12). The goal of this procedure is  two-fold, first to implement a self-learning strategy and then to acquire more and better pseudo-bitext pairs.

Unlike previous work on improving zero-shot translation~\cite{lakew2017improving,gu2019improved}, we focus on learning only a model for the {\sc zsp} languages. However, for a better analysis and fair comparison with multilingual supervised approaches, we further show  {\sc zNMT} performance by co-learning with 
language pair\fix{s} with parallel data (\textit{such as incorporating  the $S-T$ pair while learning the zero-shot $U-T$}).

An important aspect for {\sc zNMT} is how close is $U$ to $S$ in terms of vocabulary and sentence structure. Our intuition is that \mn{the closer the two languages, the higher is the performance achieved by the {\sc zNMT}. }
This will be explored in \mn{$\S$\ref{sec:results}.}

\section{Experimental Settings}
\label{sec:exp-settings}
To build an experimental ground that \fix{defines zero-shot translation without the pivot language,} we selected a real-world low-resource languages benchmark. 
In other words, \fix{we considered the data to} incorporate multiple and diverse languages, including parallel data for building strong baselines and monolingual data to evaluate {\sc zNMT}. Moreover, our choice is motivated by \mn{the} findings of~\citet{neubig18:rapid} and \citet{guzman2019two}, showing that monolingual-based approaches under-perform when assessed with real-world 
\mn{zero-shot pairs} ({\sc zsp}s).

\subsection{Languages and Dataset}
\begin{table*}[!t]
    \small
    \footnotesize
    \centering
    \resizebox{\textwidth}{!}{%
    \begin{tabular}{lcccccc}
            &           & Domain    &Az-En      &Be-En              &Gl-En      &Sk-En \\ \cline{3-7} 
        \multirow{5}{*}{Sample Size} 
                & \multirow{3}{*}{Parallel (Train/Dev/Test)}      
                    &{\sc ind}  &5.9k        &4.5K           &10.0k      &61.5k       \\
                &   &{\sc ind}          &671         &248            &682        &2271         \\
                &   &{\sc ind}          &903         &664            &1007       &2445         \\ \cline{2-7}
                & \multirow{2}{*}{Monolingual} 
                    &\textsc{ind} &5.9k, 174k   &4.5k, 201k     &10K, 174K   &61.5k, 58.6      \\ \hspace{2em}
                &   &\textsc{ood} &1.85M, 2M    &1.67M, 2M      &1.9M, 2M    &1.8M, 2M  \\ \cline{2-7}
        \multirow{1}{*}{$U$ Language Property}
                &Family/Script  &&Turkic/Latin       &Slavic/Cyrillic  &Romance/Latin    &Slavic/Latin       \\
                \bottomrule 
    \end{tabular}%
    }
    \caption{Languages and data statistics for parallel in-domain ({\sc ind}) {\sc lrp}, monolingual {\sc ind} and out-domain ({\sc ood}). 
    }
    \label{tab:langs-data}
\end{table*}

\mn{Due to their low-resource nature, we use  Ted talks data~\cite{cettolo2012wit3,qi2018andPre-trainedEmbd} for Azerbaijan\fix{i} ($Az$), Belarussian ($Be$), Galician ($Gl$), and Slovak ($Sk$)  paired with English ($En$).} The four pairs come with  train, dev, and test sets, with a max of $61k$ and as few as $4.5k$ examples, creating an ideal scenario of \fix{low-resource pair} ({\bf {\sc lrp}}). The parallel data of the {\sc lrp} is used to build baseline models in isolation and in a multilingual settings. The same dataset has been also used in recent works in an extreme\mn{ly} low-resource scenario~\cite{neubig18:rapid,xia2019generalized,lakew2019adapting}.

For the approaches utilizing monolingual data, we take the non--$En$ side for each of the four {\sc lrp} languages as in-domain ({\sc ind}) monolingual examples. For the $En$ monolingual data, segments are collected from the target side of the respective $S - T (En)$ pairs. However, to avoid the presence of comparable sentences in the $U$ and $T$ sides of the {\sc zsp}, we discard segments of monolingual $En$ if the $T (En)$ side of the $U-T$ and $S-T$ are overlapping.

For out-of-domain (\textsc{ood}) monolingual data we extract segments from Wikipedia dumps, similar to~\citet{xia2019generalized}.\footnote{Wikipedia: \href{https://dumps.wikimedia.org/}{https://dumps.wikimedia.org/}}
The collected data are de-duplicated and overlapping segments with the {\sc ind} monolingual  are removed. 
To create a practical real-world scenario that represents most of the {\sc zrp} languages, we take only the top 2$\times$10$^6$ segments, aligning with the maximum number of samples that are available for the non-$En$ languages in this benchmark. Statistics about the data are shown in Table~\ref{tab:langs-data}.

\subsection{Models}\label{subsec:models}
To test the {\sc zNMT} strategy and compare it against other approaches, we train the following models:

\begin{itemize}
\item 
\textsc{\textbf{nMT}}: trained with {\it supervised} objective using parallel \textsc{ind} data of each {\sc lrp}. 
 
\item 
\textbf{sNMT} \fix{(semi-supervised NMT)}: trained with {\it semi-supervised} objective~\cite{sennrich2015improvingMono} with back-translation leveraging {\sc ood} monolingual data. 

\item   
\textsc{\textbf{ mNMT}} \fix{(multilingual NMT)}: trained with {\it supervised} objective aggregating $116$ directions {\sc ind} parallel data~\cite{johnson2016google}.\footnote{List of languages can be found in the Appendix.} 

\item    
{\bf uNMT} \fix{(unsupervised NMT)}: trained with {\it unsupervised} objectives~\cite{lample2019cross} leveraging \textsc{ind} and \textsc{ood} data. 
\item    
{\bf zNMT} \fix{(zeros-shot NMT)}: trained with proposed {\it zero-shot} modeling as in Algorithm~\ref{algo:zNMT} leveraging \textsc{ind} and \textsc{ood} monolingual data.
\end{itemize}

\subsection{Pre-Training Objectives}\label{subsec:pre-training-obj}
\textsc{uNMT} leverages a cross-lingual masked language model pre-training ({\sc mLM}). We train the \textsc{mLM} following the~\cite{lample2019cross} settings, using both \textsc{ood} and \textsc{ind} monolingual data of each \textsc{zsp} language. 
Although \textsc{zNMT} can be initialized from any pre-trained NMT model as long as the $T$ language of the {\sc zsp} is observed (see $\S$\ref{sec:znmt}), we devise three types of pre-training strategies for a rigorous evaluation and based on data availability:

\begin{itemize}
    \item 
    {\sc biTM} -- \fix{four} bilingual 
    \fix{translation models} trained with~$S\leftrightarrow En$ parallel data.\footnote{$S$ is a 
related language to the $unseen$-$U$. The $S/unseen$-$U$ combinations are: $Az$/Turkish($Tr$), $Be$/Russian($Ru$), $Gl$/Portuguese($Pt$), and $Sk$/Czech($Cs$).}

    \item 
    {\sc muTM}$100$ -- a multilingual NMT model with \fix{$100$ translation directions from the TED talks data,} \fix{excluding} the four {\sc zsp} and the pairs used for {\sc bi\fix{T}M}. 

    \item 
    {\sc mu\fix{T}M}$108$ -- 
    a \fix{similar} multilingual model \fix{with {\sc muTM}$100$, however,} including \fix{additional 8 directions} 
    used for the {\sc bi\fix{T}M} \fix{models}.
\end{itemize}

The idea behind the \fix{{\sc muTM}$100$} and \fix{{\sc muTM}$108$} strategies is to check \mn{to what extent} the presence of close languages to the $unseen$-$U$ in the pre-trained model can \mn{support}
the \textsc{zNMT} approach. Note \mn{that}, unlike in the {\sc mLM}, all the pre-training for \textsc{zNMT} utilized only \fix{in-domain}\footnote{Utilizing \textsc{ood} monolingual data for NMT pre-training could be an advantageous and interesting direction to investigate, however, for this work we constrain to utilizing only \textsc{ind} data.}

\begin{table*}[!ht]
\small
\resizebox{\textwidth}{!}{%
\begin{tabular}{clcl|ll|ll|ll|ll}
\toprule
                    
Id &Model          &Pre-Train   &Scen.      &Az-En  &En-Az  &Be-En  &En-Be &Gl-En  &En-Gl   &Sk-En &En-Sk \\ \midrule 
1& Supervised ({\sc nMT})          
    &-  &\textsc{ind} &3.60 &2.07   &5.20  &3.40  &19.53   &15.52 &27.24   &20.91 \\ \cline{4-12}
2& Semi-Supervised ({\sc sNMT})     
    &- &\textsc{iod} &3.74   &1.92   &5.74   &4.03   &22.08  &17.27 &{\bf 27.85}       &{\bf 21.24} \\ \midrule 
3&\multirow{4}{*}{Unsupervised ({\sc uNMT})} 
&\multirow{4}{*}{\sc mLM} 
    &{\sc ind}   &1.97   &1.56   &4.61   &1.47   &13.93  &5.89   &15.70  &11.91 \\
4&    &&{\sc ood}   &3.26   &2.55   &5.69   &3.73   &16.71  &14.90  &10.62 &7.62      \\
5&    &&{\sc i-od}   &0.88   &1.18   &0.82   &0.90   &5.06   &2.78   &6.39   &7.28  \\ \cline{4-12}
6&    && {\sc iod} &3.97   &2.57   &5.57   &3.78   &20.23  &17.07  &13.77  &11.43 \\ \midrule  
7&\multirow{4}{*}{Zero-Shot ({\sc zNMT})}  
    &\multirow{4}{*}{\sc biTM} 
        &{\sc ind}      &8.86   &4.87   &4.42       &3.45   &23.57  &18.17  &17.89 &14.08 \\
8&        &&{\sc ood}     &6.76   &4.45   &5.75       &5.16   &17.28  &16.97  &9.13  &6.74 \\
9&        &&{\sc i-od}    &2.63   &3.96   &1.20       &2.23   &14.96   &16.23 &9.10 &11.35 \\\cline{4-12}
$10$&        &&{\sc iod}     &{\bf11.38}   &{\bf6.28}     &{\bf7.36}   &{\bf6.35}   &\textbf{25.46}
        &\textbf{21.09} &19.43 &14.70 \\
        \bottomrule

\end{tabular}%
}
\caption{Results from low-resource supervised and semi-supervised, and our monolingual based {\sc zNMT} in comparison with {\sc uNMT}~\cite{lample2019cross} across the four training scenarios.} 
\label{tab:biPM-results}
\end{table*}

\subsection{Training Scenarios} 
\label{subsec:training-criterion-comp}
We define \mn{four model training criteria}
based on a real-world scenario for a {\sc zsp}, that is the availability and characteristics (\textit{such as domain and size}) of monolingual data. 

\begin{itemize}
\item {\bf IND}: in-domain data is \mn{used} both on the $U$ and $T$ zero-shot sides.

\item {\bf OOD}: out-of-domain data are used  both in the $U$ and $T$ sides of the {\sc zsp}.

\item {\bf I-OD}: \mn{a}  scenario where we create a domain mismatch between the $U$ and $T$ side of the {\sc zsp}, by replacing the $T$ {\sc ind} with {\sc ood} data.

\item {\bf IOD}: \mn{a} the mix of {\sc ind} and relatively large {\sc ood} data \mn{is} used on both $U$ and $T$ sides.

\end{itemize}

\subsection{Training Pipeline} 
\textbf{Data Preparation:} 
we collect the {\sc ind} Ted talks data provided by~\citet{qi2018andPre-trainedEmbd} and {\sc ood} Wikipedia\footnote{WikiExtractor: \href{https://github.com/attardi/wikiextractor}{https://github.com/attardi/wikiextractor}} data,  and then segment them into sub-word units. We use SentencePiece~\citep{kudo2018sentencepiece}\footnote{SentencePiece: \href{https://github.com/google/sentencepiece}{https://github.com/google/sentencepiece}} to learn BPE with $32$k merge operations using the {\sc ind} training data, whereas for {
\sc uNMT} we also use {\sc ood} monolingual data.
\\

\noindent
{\bf Model Settings:} all experiments use Transformer~\cite{vaswani2017attention}. 
{\sc uNMT} is trained using the XLM
tool~\cite{lample2019cross}\footnote{XLM: \href{https://github.com/facebookresearch/XLM}{
https://github.com/facebookresearch/XLM}},
while for the rest we utilize OpenNMT~\cite{klein2017opennmt}.\footnote{OpenNMT: \href{https://github.com/OpenNMT}{https://github.com/OpenNMT}}
Models are configured with $512$ dimension, $8$ headed $6$ self-attention layers, and $2048$ feed-forward dimension. Additional configuration details are provided in the Appendix.
\\

\noindent
\textbf{Evaluation:} we use the BLEU score ~\citep{papineni2002bleu}\footnote{Moses Toolkit: \href{http://www.statmt.org/moses}{http://www.statmt.org/moses}} for assessing models' performance.
Scores are computed on detokenized (\textit{hypothesis}, \textit{reference}) pairs. The checkpoints with best BLEU on \mn{the} dev set are used for the final evaluations.

\section{Results and Analysis}
\label{sec:results}
In Table~\ref{tab:biPM-results}, we asses the quality of various NMT systems featuring different model types ($\S$\ref{subsec:models}), training 
scenarios ($\S$\ref{subsec:training-criterion-comp}) using  bilingual {\sc biTM} for {\sc zNMT} and {\sc mLM} for {\sc uNMT} pre-training.
We then show the effect of leveraging massive multilingual pre-training on the {\sc zNMT} performance (Table~\ref{tab:muPM-results}). Finally, we expand our analysis to co-learning {\sc zNMT} with supervised NMT (Table~\ref{tab:muPM-colearning-results}). \mn{A} preliminary assessment \mn{of the experimental choices adopted for}  {\sc zNMT} can be found in the Appendix.

\begin{table*}[!t]
\resizebox{\textwidth}{!}{%
\begin{tabular}{clcl|ll|ll|ll|ll}
\toprule
Id& Model      &Pre-Train   &Scen.  &Az-En  &En-Az  &Be-En  &En-Be &Gl-En   &En-Gl &Sk-En & En-Sk \\ \midrule 
1 &{Supervised ({\sc mNMT})}        
&-&{\sc ind}  &\textbf{11.37} &\textbf{4.98}    &\textbf{18.36}       &10.06       &29.77   &25.44   &27.49   &\textbf{22.72} \\ \midrule

2& \multirow{4}{*}{Zero-Shot ({\sc zNMT})}
                &\multirow{2}{*}{\sc muTM100}
                    &{\sc i-od}    &1.04   &2.55   &7.31   &7.14   &22.91 &22.93  &12.33 &11.81 \\ 
3&                    &&{\sc iod}   &2.51   &1.55   &16.20  &10.30  &32.14 &26.68  &23.52   &16.60 \\ \cline{3-12} 
                    
4&  
                &\multirow{2}{*}{\sc muTM108}
                    &{\sc i-od}   &4.14   &2.38   &10.18  &9.00       &26.45       &25.34       &20.26  &19.69 \\  
5&                    &&{\sc iod}   & 9.19   &2.75   &17.26  &\textbf{10.95}  &\textbf{32.83}   &\textbf{27.49}  &\textbf{28.94}       &21.53 \\ 
                    \bottomrule 
\end{tabular}%
}
\caption{{\sc zNMT} when initialized from multilingual pre-trained models, in comparison with supervised {\sc mNMT}. 
} 
\label{tab:muPM-results}
\end{table*}

\begin{figure}[!t]
    \centering
    \includegraphics[width=\linewidth]{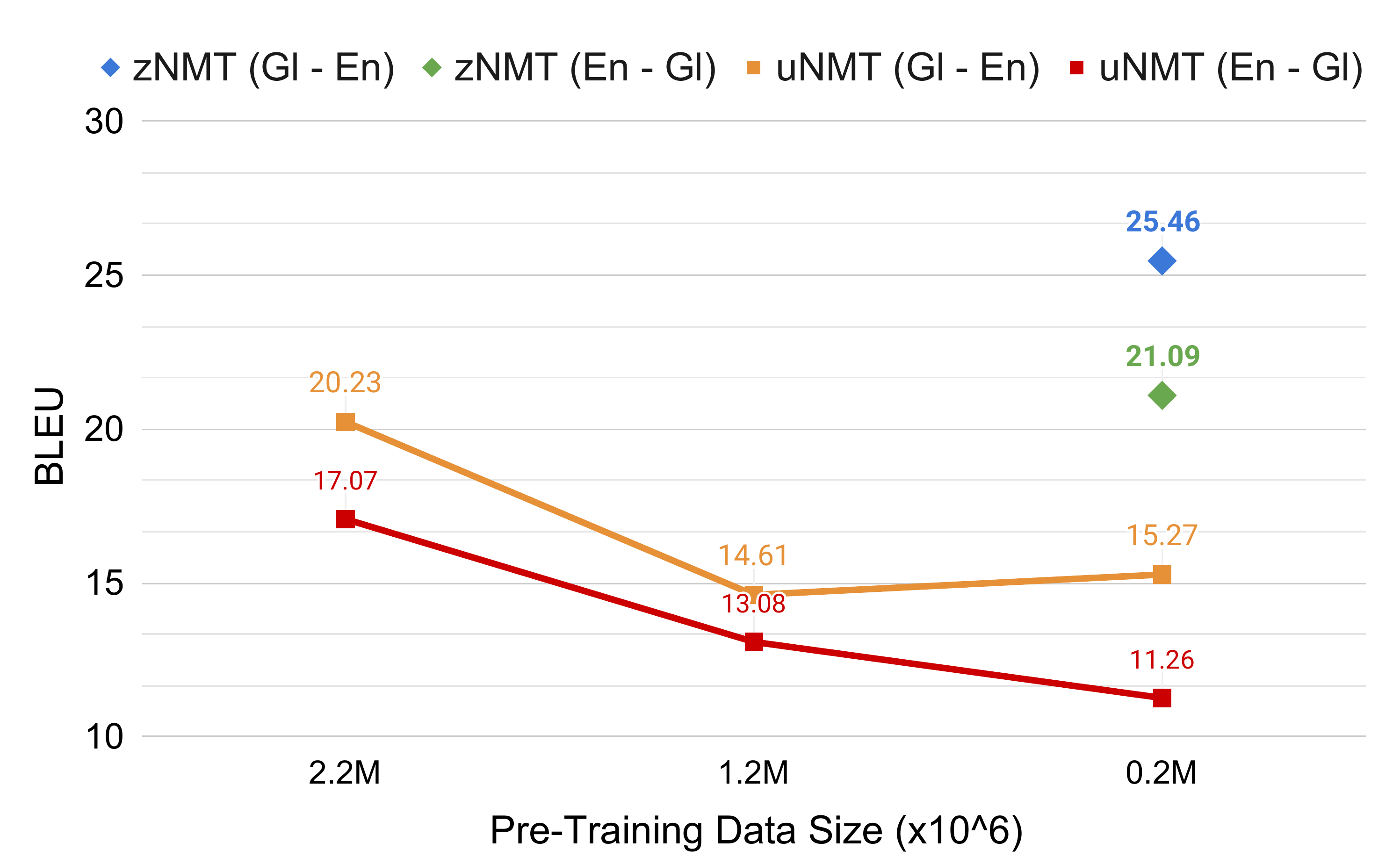}
    \caption{Effect of pre-training data size.} 
    \label{fig:zNMT-vs-uNMT-diff-pretraining}
\end{figure}

\subsection{Bilingual Pre-Training}\label{subsec:zNMT-from-biPM}
The first two rows of Table~\ref{tab:biPM-results} confirm the results of~\cite{sennrich-etal-2016-improving} showing that semi-supervised approaches, which leverage back-translation, outperform supervised NMT systems. Moreover, the performance of both approaches strongly relates to the quantity of the available training data ($Gl-En \gg Be-En$).

In the {\bf in-domain} training scenario ({\sc ind}), our {\sc zNMT} approach outperforms the  supervised low-resource NMT, 
except for $Sk\leftrightarrow En$ and $Be\rightarrow En$ ($rows$ 1, 7), demonstrating the effectiveness of our proposal in leveraging monolingual data. 
The advantage of {\sc zNMT} is further confirmed when comparing it with {\sc uNMT}. In this case,  {\sc zNMT} outperforms  {\sc uNMT} in \mn{7 out of 8 language} directions and it is on par on the $Be\rightarrow En$ language pair.

In the {\bf out-of-domain} training scenario ({\sc ood}), 
despite the fact that {\sc uNMT} utilizes $\times10$ more {\sc ood} segments than {\sc zNMT}, our approach 
surprisingly achieves better performance than {\sc uNMT}, except for $Sk-En$ (\textit{rows} 4, 8). 
Fig.~\ref{fig:zNMT-vs-uNMT-diff-pretraining} shows the effect of varying the amount of monolingual data during pre-training ({\sc biTM, mLM}). We observe that {\sc uNMT} is significantly 
affected by decreasing the size of the monolingual data 
and, when using the same quantity applied in {\sc zNMT} (200k), it achieves much worse performance (-10 BLEU points).  
Our findings clearly show the effectiveness of {\sc zNMT} in learning better with small monolingual data, a case applicable for most {\sc lrp} and {\sc zsp}.

The {\bf domain mismatch} scenario ({\sc i-od}) 
is the most realistic representation \mn{for {\sc zsp} and {\sc lrp} settings, as it does not count on access} to comparable monolingual data. Both {\sc zNMT} and {\sc uNMT} show drastic \mn{performance drops}
in all directions (\textit{rows} 5, 9), confirming \mn{the findings of} \citet{kim2020and}. Besides the performance drop, {\sc zNMT} shows \mn{higher robustness to domain shifts, resulting in higher scores.} \mn{ {\sc uNMT}, in contrast, is susceptible} to the domain divergence and requires comparable monolingual data that is hard to acquire for {\sc zrp}.

In the \textbf{mixed domain} scenario ({\sc iod}), {\sc zNMT} 
prevails over {\sc uNMT} by a larger margin (\textit{rows} 6, 10) ranging from $+$1.79 ($Az-En$) to +7.41 ($Be-En$) BLEU. 
This is a similar trend to the {\sc ind, ood}, and {\sc i-od} scenarios, 
validating the superiority of the \mn{proposed} {\sc zNMT} learning approach. 
\mn{A more interesting aspect is that, except for $Sk-En$ (\textit{rows} 2, 10),  {\sc zNMT} also outperforms semi-supervised NMT.}
This shows that \mn{a stronger}
model can be learned exploiting as few as $200k$ monolingual data with our {\sc zNMT} learning principles, 
\fix{in comparison with a {\sc lrp} performance (such as $Az-En$ with $5.9k$, and $Gl-En$ with 10k parallel data)}.

In sum, the results in Table~\ref{tab:biPM-results}  \mn{show that, for low-resource language pairs, {\sc zNMT} leveraging {\sc biTM} can in most of the cases outperform  supervised NMT  trained on language-specific parallel data. }
Moreover, {\sc zNMT} is robust towards domain shifts from the pre-training and across $U - T$ {\sc zsp}, outperforming unsupervised NMT in all training 
scenarios.

\begin{table*}[!t]
\resizebox{\textwidth}{!}{%
\begin{tabular}{clcl|ll|ll|ll|ll}
\toprule
Id& Model      &Pre-Train   &Scen.  &Az-En  &En-Az  &Be-En  &En-Be &Gl-En   &En-Gl &Sk-En & En-Sk \\ \midrule

1   &{\sc massive} 
&-         &{\sc ind}    &12.78  &5.06       &21.73  &\textbf{10.72}  &30.65  &26.59       &29.54 &24.52 \\
2   &{\sc dynAdapt} 
&{\sc muTM108}         &{\sc ind}    &15.33  &-       &23.80  &-       &\textbf{34.18}  &-       &\textbf{32.48}&- \\ \cline{4-12}
3   &{\sc augAdapt} 
&{\sc muTM108}           &{\sc iod}  &\textbf{15.74}  &-       &\textbf{24.51}  &-       &33.16  &-       &32.17 &- \\ \midrule 
                    
4&   \multirow{2}{*}{{\sc zNMT + co-Learning}}
    &\multirow{2}{*}{\sc muTM108}
        &{\sc ind}   &8.56   &2.25    &16.34  &9.16  &32.75   &26.75  &28.84 &22.34 \\  
5&        &&{\sc iod}  &11.01   &2.28      &18.05  & 10.38       &33.26     &\textbf{27.64}       &29.94     &22.26 \\ \bottomrule
\end{tabular}%
}
\caption{Performance of {\sc zNMT} with co-Learning 
in comparison 
with the supervised {\sc massive}~\cite{aharoni2019massively} and {\sc dynAdapt}~\cite{lakew2019adapting}, and semi-supervised {\sc augAdapt}~\cite{xia2019generalized}. 
} 
\label{tab:muPM-colearning-results}
\end{table*}

\subsection{Multilingual Pre-Training}
\label{subsec:zNMT-from-muPM}
To test the capability of {\sc zNMT} \mn{to leverage}
universal representation from the pre-trained model,
we built two massive multilingual systems: {\sc muTM100} that excludes the pairs used for {\sc biTM} and {\sc muTM108} that assumes a favorable condition by also including the {\sc biTM} pairs.

Besides the initialization from the universal models, we train the best {\sc zNMT} scenario ({\sc iod}) and the most challenging one ({\sc i-od}) from~Table~\ref{tab:biPM-results} by only using the $U-T$ monolingual data.
Table~\ref{tab:muPM-results} (\textit{row} 1) shows that the supervised {\sc mNMT} benefits from the multilingual corpus (i.e., \fix{trained with 116} \fix{directions data including the zero-shot pairs}), and, as expected, obtains improvements over the bilingual supervised models in Table~\ref{tab:biPM-results}.

In the {\bf domain mismatch} scenario ({\sc i-od}), the use of {\sc biTM} leads to large drops in performance compared to the {\sc ind} or {\sc iod} scenario (Table~\ref{tab:biPM-results}, \textit{row} 9). 
This is also confirmed when leveraging the {\sc muTM*} pre-training (\textit{rows} 2, 4). 
However, the robust multilingual pre-training shows improvements compared with the initialization from {\sc biTM}. 
For instance, the $Gl\rightarrow En$ with {\sc biTM} drops -$10.5$ (from 25.46 with {\sc iod} to 14.96 with {\sc i-od}), while {\sc mu\fix{T}M108} degrades only by $6.38$ BLEU points.

Our approach leveraging the {\bf mixed domain} ({\sc iod}) monolingual data with {\sc muTM108} achieves the best performance in most of the language directions and is on par or even better with the supervised multilingual (\textit{rows} 1, 3, and 5). This is a remarkable result because the {\sc zNMT} systems do not leverage any language-specific parallel data.

The advantage of using a {\bf robust pre-training} can be 
ascribed to the availability of multiple languages that  maximizes the lexical and linguistic similarity with the {\sc zsp}. Looking at the {\sc iod} scenario {\sc muTM*} in Table~\ref{tab:muPM-results} (\textit{rows}~3,~5), we notice an overall improvement over {\sc biTM} pre-training (Table~\ref{tab:biPM-results}, \textit{row} 10). A comparison against the best supervised {\sc sNMT} model (Table~\ref{tab:biPM-results}) using low-resource parallel data shows better performance of {\sc zNMT} with {\sc muTM108} up to (+10.75 $\leftrightarrow$ +10.22) for $Gl-En$.
However, as for the 
{\sc biTM}, it is not always the case to find closely related $S-T$ pair(s) to the $U-T$ {\sc zsp} for pre-training.
Hence, 
it is rather more interesting to observe that {\sc zNMT} can learn even better with {\sc muTM100} 
without observing the most related languages as in {\sc biTM}.

With respect to the {\sc biTM}, $Az-En$ is the only {\sc zsp} that do not benefit from the multilingual pre-training. \mn{One possible reason is the absence of related language pairs, which makes the pre-training representation  dominated by other pairs.}
This becomes more evident 
(lower BLEU) when using {\sc muTM100}, a pre-training that excludes the closest $S-T$ pair to $Az-En$.

In sum, our approach shows significant improvements when leveraging universal pre-trained models. This is demonstrated by the large gains in performance in all the 
scenarios over the {\sc biTM} pre-training. The fact that our method is able to approach the performance of the multilingual supervised settings, and in some cases to overcome them, makes it a valuable solution for {\sc zsp} languages.

\subsection{Co-Learning with Supervised Directions}
To test the complementary of {\sc zNMT} and supervised NMT, we add the parallel data of the latter only at the learning stage. Although it is possible to leverage multilingual parallel data, in this experiments we only utilize a single $S-T$ parallel pair from the {\sc biTM} for the zero-shot co-learning stage of $U-T$.

We compare our co-learning system with three state-of-the-art approaches:~{\sc massive} trains a many-to-many system on all ($116\leftrightarrow 116$) available pairs~\cite{aharoni2019massively},~\textsc{dynAdapt}~\cite{lakew2019adapting} uses an {\sc ind} criterion to adapt {\sc mu\fix{T}M108} pre-trained model by first tailoring the vocabulary and embeddings to the {\sc lrp} and~\textsc{augAdapt}~\cite{xia2019generalized} generates psudo-bitext from {\sc ood} monolingual and adapts {\sc muTM108} together with the {\sc iod} data.  
The latter two utilize a similar co-learning strategy during the adaptation of the universal model with the 
parallel data, and reported results only when the target is $En$. Similar to {\sc massive}, {\sc dynAdapt}, and {\sc augAdapt}, 
we focused on {\sc ind} and {\sc iod} training scenarios. 

Table~\ref{tab:muPM-colearning-results} 
reports the performance of these approaches and of {\sc zNMT} with co-learning using in the {\sc ind} and {\sc iod} 
scenarios. 
Comparing {\sc zNMT} + {\sc co-learning} (\textit{rows} 5) with {\sc zNMT} in Table~\ref{tab:muPM-results} (\textit{row} 5), the results show that \textit{co-learning} generally leads to better performance. 
However, when the target language is non-$En$, the differences are marginal and the two approaches can be considered comparable. 
This is directly associated with the fact that we have more $En$ segments, from the aggregation of the $S-T(En)$ and $U-T(En)$ pairs.    
\textsc{dynAdapt} and \textsc{augAdapt} are the two best performing supervised techniques on the this benchmark, but {\sc zNMT} with co-learning achieves competitive performance approaching them both in the {\sc ind} and {\sc iod} scenarios.

Overall,  these findings show that our approach makes it possible to extend zero-shot NMT to an unseen language $U$. In particular, leveraging a universal pre-training model and co-learning with supervised task allows our approach to learn a better NMT model from monolingual data.

\section{Conclusion} 
\mn{We presented} a new zero-shot NMT modeling variant, specifically 
\mn{targeting}
languages that have never been observed in a pre-trained NMT. We \mn{showed}
limitations of current approaches with the pivot language premise and zero-shot translation only between observed languages, and \mn{proposed}
a relaxation to zero-shot NMT to incorporate  
unseen language\fix{s}. Our approach includes initialization, augmentation, and training stages to construct
a self-learning cycle to incrementally correct the primal and dual zero-shot translation quality. We empirically \mn{demonstrated}
the effectiveness of the proposed approach using diverse real-world zero-resource languages in in-domain, out-of-domain, domain-mismatch, and mixed domain scenarios. Results both from bilingual and multilingual initialization \mn{ not only revealed}
the possibility of extending zero-shot NMT for unseen languages but also improved performance over unsupervised, low-resource supervised and semi-supervised NMT.


\bibliography{anthology,eacl2021}
\bibliographystyle{acl_natbib}

\clearpage
\appendix

\section{Preliminary Assessment}\label{sec:preliminary-assessment} 
We summarize the motivation for certain experimental design choices in 
our zero-shot NMT ({\sc zNMT}) modeling, 
analyzing model pre-training type (such as bilingual ({\sc biTM}) and multilingual ({\sc muTM*})) 
and effective utilization of the in-domain ({\sc ind}), out-of-domain ({\sc ood}), mixed domain ({\sc iod}) 
monolingual data. The $Gl-En$ zero-shot pair ({\sc zsp}) is used for our assessment.

\begin{figure}[hbt!]
    \includegraphics[width=\linewidth]{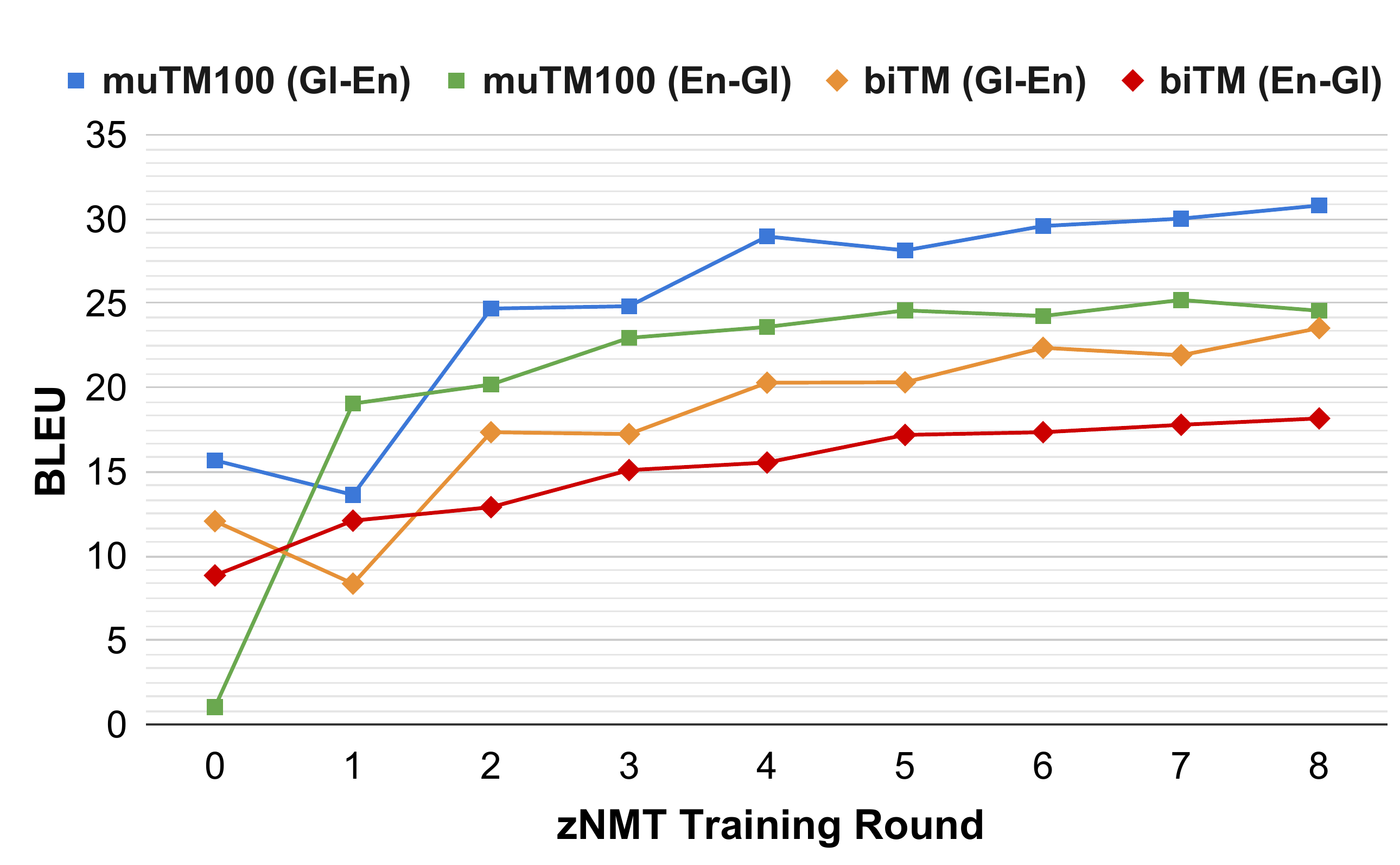}
    \caption{Performance of {\sc zNMT} using bilingual ({\sc biTM}) and multilingual ({\sc muTM100}) pre-trainings.}
    \label{fig:preliminary-assessment-1}
\end{figure}

\noindent
{\bf Pre-Trained NMT Variant} 

\noindent
Unlike previous work in zero-shot NMT, 
our {\sc zNMT} aims to leverage both bilingual and multilingual pre-trainings.  Fig.~\ref{fig:preliminary-assessment-1} 
shows {\sc zNMT} improves better if initialized from multilingual pre-training ({\sc muTM100}). 
This is despite {\sc muTM100} not observing the closest language pair ($Pt-En$) to the {\sc zsp} ($Gl-En$), while {\sc biTM} is trained using only $Pt-En$.  
Hence, the gain by initializing from {\sc muTM100} shows the robustness of pre-training with multiple languages and its positive effect on {\sc zNMT}. 
However, these results signal {\sc muTM*} importance for {\sc zNMT} modeling, for further verification and better comparison with the bilingual supervised and unsupervised approaches our main experimental setup first focuses on utilizing {\sc biTM}.

\begin{figure}[hbt!]
    \includegraphics[width=\linewidth]{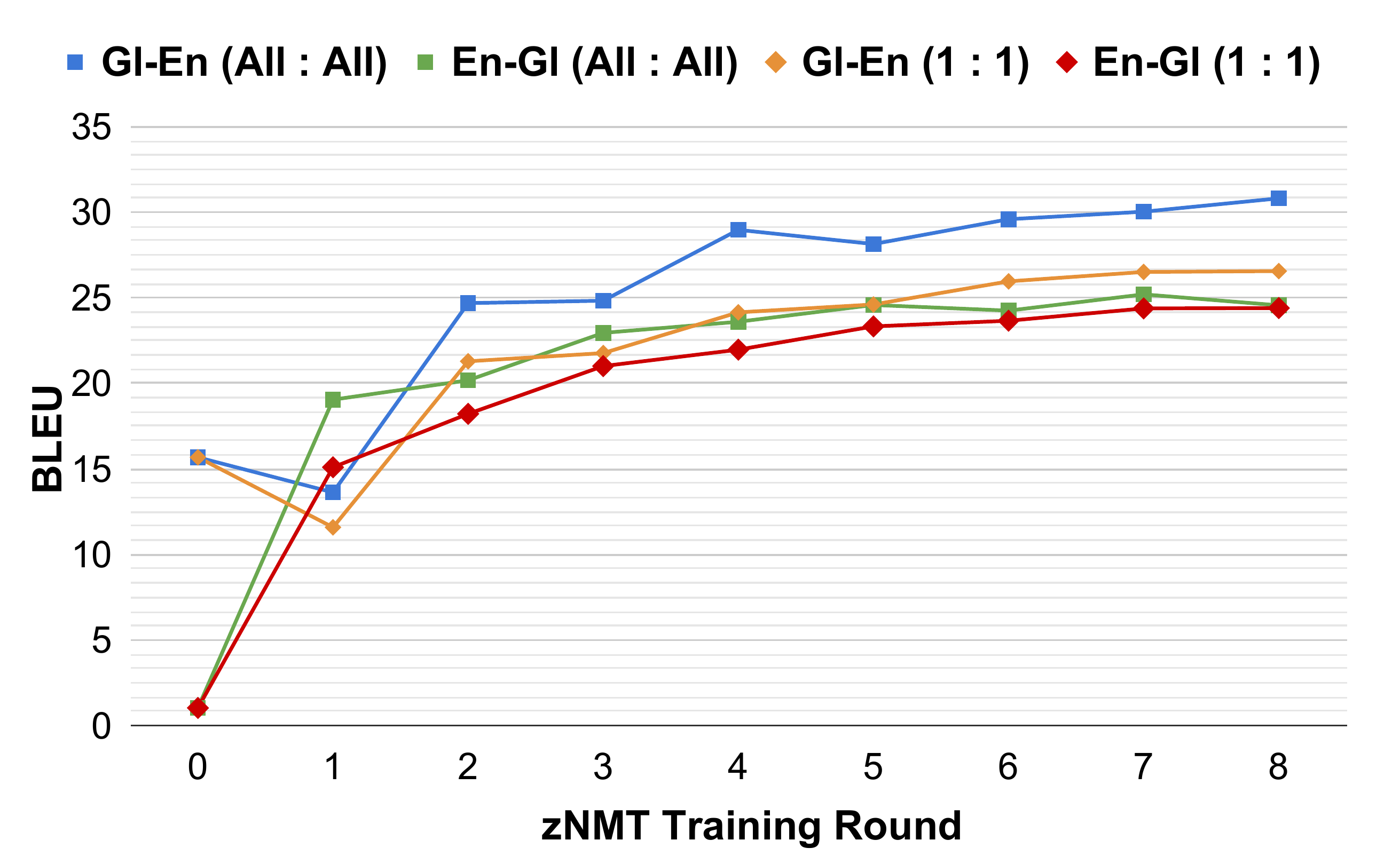}
    \caption{Performance of {\sc zNMT} by varying monolingual data size ratio between  $U$ and $T$.}
    \label{fig:preliminary-assessment-2}
\end{figure}

\vspace{1em}
\noindent
{\bf Data Size and Domain} 

\noindent
For the training scenarios involving {\sc ind} and {\sc ood} data, 
Fig.~\ref{fig:preliminary-assessment-2} 
shows if available using all {\sc ind} segments (All:All) is better than taking equal proportion (1:1) of the $U$ and $T$ sides of the {\sc zsp}. 
In a parallel 
experiment for {\sc iod} scenario, however, we observed that balancing the {\sc ood} segments with {\sc ind} lead to a comparable or better performance. 
In other words, we select {\sc ood} proportionally ($\approx 200k$) to the largest {\sc ind} side of the {\sc zsp}. 
We noted a similar trend 
for semi-supervised (\textsc{sNMT}) low-resource model, that shows better performance when using $\approx 200k$ {\sc ood} leading to 
$22.08\leftrightarrow 17.27$ ($Gl\leftrightarrow En$) 
than using $\approx2M$ segments that degrades to $20.86\leftrightarrow 16.44$ BLEU.
However, for {\sc uNMT} it is a common knowledge where more monolingual data leads to better performance~\cite{lample2019cross}. 
We confirmed this by reducing the {\sc ood} to $200k$ as in {\sc zNMT} and {\sc sNMT}, where we observed a $5$ BLEU drop in {\sc uNMT} performance in both $Gl\leftrightarrow En$ directions. 
For this reason, we train {\sc uNMT} models using all the available {\sc ind} and $\approx 2M$ {\sc ood} segments. 
In other words, the unsupervised models 
consume all the available {\sc ind} and {\sc ood} monolingual data, that is $\times10$ more than the {\sc sNMT} baseline and our {\sc zNMT} utilized. 
In sum, this shows the efficiency of our approach to reach to a better performance with less resources. Detail comparisons are provided in the main experimental section. 

\begin{figure}[hbt!]
    \includegraphics[width=\linewidth]{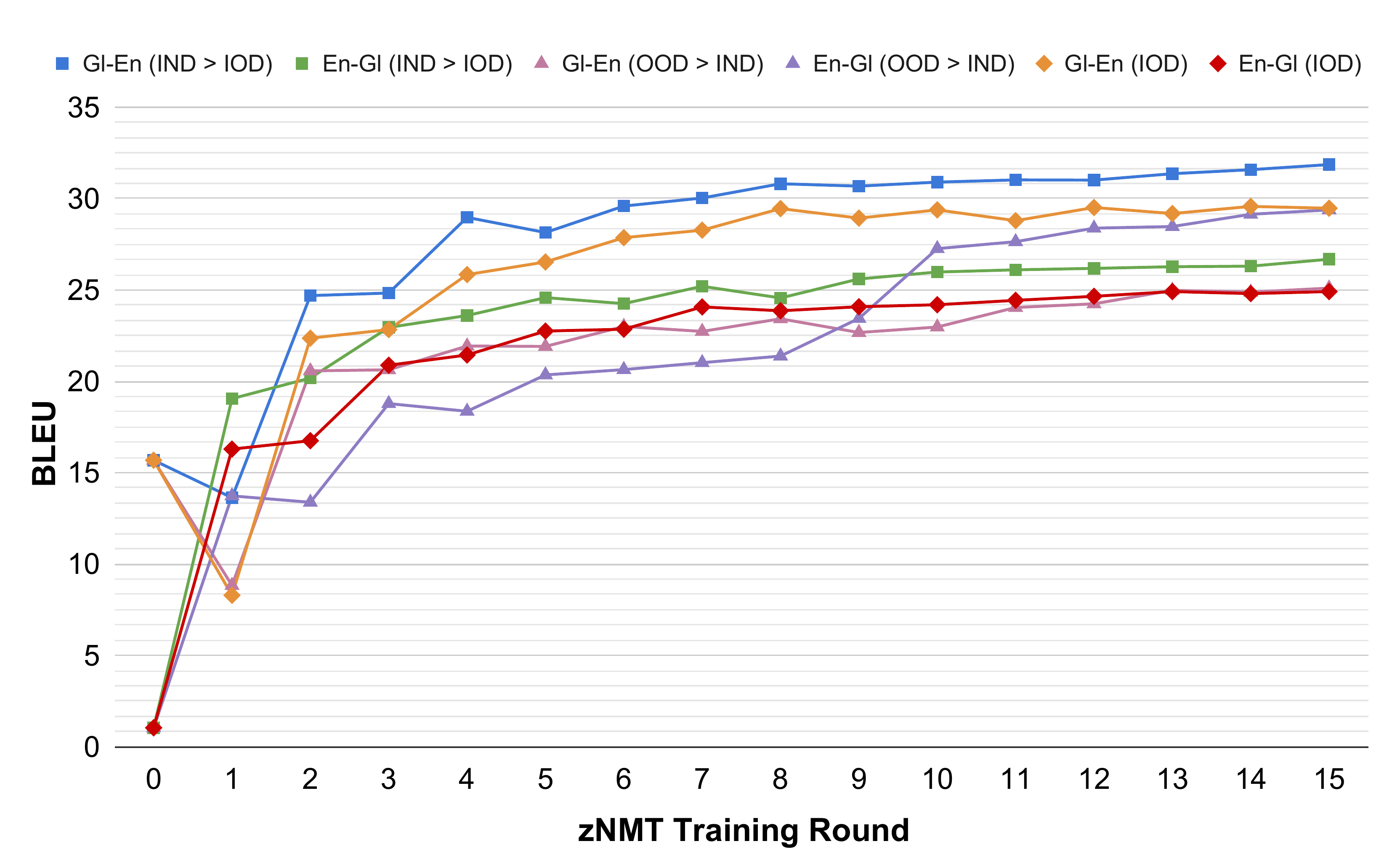}
    \caption{Training strategies to best utilize in-domain ({\sc ind}) and out-of-domain ({\sc ood}) monolingual data.}
    \label{fig:preliminary-assessment-3}
\end{figure}

Lastly, Fig.~\ref{fig:preliminary-assessment-3} 
shows 
an effective strategy of utilizing the {\sc ind} and {\sc ood} data for \textsc{zNMT} in a mixed domain ({\sc iod}) scenario. 
The test settings show first learning {\sc zNMT} with the {\sc ind} and progressively incorporating {\sc ood} data ({\sc ind $>$ iod}) is the best 
approach, in comparison with 
({\sc ood $>$ iod}), or utilizing 
({\sc iod}) from the beginning. 
Considering 
pre-trained models for {\sc zNMT} utilizes 
{\sc ind} data (except for the unseen $U$), the finding is 
expected and leads to a better performance.
Applying a similar ({\sc ind $>$ iod}) strategy for \textsc{uNMT}, however, resulted in a drop of up to $7$ BLEU for $En\rightarrow Gl$, compared to training with the mixed domain ({\sc iod}) from the beginning. 
This is likely due to the fact that the 
the pre-training for {\sc uNMT} observes both {\sc id} and {\sc od} data of $U-T$ {\sc zsp} and 
leading to a better learning when using {\sc iod}. 
In our main experimental setup we choose the best strategy for each of the approaches.

\section{Model Configuration and Parameters}\label{app:model-config}
To tackle over-fitting in the bilingual baseline supervised and semi-supervised NMT models we employ a dropout rate of $0.1$ on the attention and $0.3$ on all the other layers.  
Whereas the dropout rate for all the other models are set uniformly to $0.1$.
We use source and target tied embeddings~\cite{press2016using}.  
Samples exceeding $100$ sub-word counts are discarded at time of training. 
Model training is done on a single V$100$ GPU with batch-size of $4,096$ tokens.
Adam is used as an optimizer~\cite{kingma2014adam} with a learning rate of $10^{-4}$.

Details about model parameter are provided in Table~\ref{tab:model-stats}. 
At time of training all models have shown to converge. While {\sc zNMT} shows the fastest learning curve within $15-20$ epochs, {\sc uNMT} run up to $100$ epochs to reach convergence.

\begin{table}[hbt!]
    \resizebox{\linewidth}{!}{%
    \begin{tabular}{c|ccc}
        \toprule
        Model       & Initialization    &Params ($\times$10$^6$) &Layers  \\   
        \hline
        {\sc mLM}   & -                 &41           &6             \\   
        {\sc biTM}  & -                 &53           &6              \\  
        {\sc muTM$^*$}  & -             &69           &6                \\ \hline 
        {\sc nMT}   & -                 &38           &4              \\ 
        {\sc sNMT}   & -                 &38           &4             \\ 
        {\sc mnMT}  & -                 &69           &6             \\ 
        \hline 
        {\sc uNMT}   & {\sc mLM}         &86           &6            \\ 
        {\sc zNMT}   & {\sc biTM}        &53           &6            \\ 
        {\sc zNMT}   & {\sc muTM$^*$}    &69           &6            \\ 
        \bottomrule
    \end{tabular}%
    }
    \caption{Model, parameter size, and number of self-attention layers. {\sc muTM*} represents both {\sc mu100} and {\sc mu108}.}
    \label{tab:model-stats}
\end{table}

\section{Languages and Data}
Table~\ref{tab:ted-talk-languages} lists the languages and examples size from the TED talks data. 

\begin{table*}[!hbt]
    \footnotesize
    \centering
    \begin{tabular}{lllll}

\toprule
Language    & Lang. Id      &Train              &Dev            &Test \\ \hline
Arabic      &   ar         &214111          & 4714  &          5953 \\
%
%
\rowcolor[gray]{0.9} Azerbaijani &az          & 5946     &       671&             903 \\
\rowcolor[gray]{0.9} Belarusian & be           &4509    &        248 &            664 \\
Bulgarian   & bg         &174444   &        4082  &          5060 \\
Bengali & bn          & 4649  &          896   &          216 \\
Bosnian & bs          & 5664 &           474    &         463 \\
Czech   &cs         &103093          & 3462     &       3831 \\
Danish  &da          &44940         &  1694      &      1683 \\
German  &de         &167888        &   4148       &     4491 \\
Greek   &el         &134327       &    3344        &    4431 \\
Esperanto   &eo          & 6535      &      495         &    758 \\
Spanish &es         &196026     &      4231          &  5571 \\
Estonian    &et          &10738    &        740           & 1087 \\
Basque  &eu           &5182   &         318 &            379 \\
Persian &fa         &150965  &         3930  &          4490 \\
Finnish &fi          &24222           & 981   &         1301 \\
French-Canadian  &fr-ca       &19870          &  838    &        1611 \\
French  &fr         &192304         &  4320     &       4866 \\
\rowcolor[gray]{0.9} Galician   &gl          &10017        &    682      &      1007 \\
Hebrew  &he         &211819       &    4515       &     5508 \\
Hindi   &hi          &18798      &      854        &    1243 \\
Croatian    &hr         &122091     &      3333         &   4881 \\
Hungarian   &hu         &147219    &       3725 &           4981 \\
Armenian    &hy          &21360   &         739  &          1567 \\
Indonesian  &id          &87406  &         2677   &         3179 \\
Italian &it         &204503 &          4547    &        5625 \\
Japanese   &ja         &204090         &  4429     &       5565 \\
Georgian    &ka          &13193        &    654      &       943 \\
Kazakh  &kk           &3317       &     938      &       775 \\
Korean  &ko         &205640      &     4441       &     5637 \\
Kurdish &ku          &10371     &       265        &     766 \\
Lithuanian  &lt          &41919    &       1791         &   1791 \\
Macedonian  &mk          &25335   &         640          &   438 \\
Mongolian   &mn           &7607  &          372           &  414 \\
Marathi &mr           &9840  &          767 &           1090 \\
Malay   &ms           &5220 &           539  &           260 \\
Burmese &my          &21497           & 741   &         1504 \\
Norwegian     &nb          &15825          &  826    &         806 \\ 
Dutch   &nl         &183767         &  4459     &       5006 \\
Polish  &pl         &176169        &   4108      &      5010 \\
Portuguese-Brazilian    &pt-br      &184755       &    4035       &     4855 \\
Portuguese  &pt          &51785      &     1193        &    1803 \\
Romanian    &ro         &180484     &      3904         &   4631 \\
Russian &ru         &208458    &       4814          &  5483 \\
\rowcolor[gray]{0.9} Slovak &sk          &61470   &        2271           & 2445 \\
Slovenian   &sl          &19831  &         1068            &1251 \\
Albanian    &sq          &44525 &          1556 &           2443 \\
Serbian &sr         &136898           &3798   &         4634 \\
Swedish &sv          &56647          & 1729    &        2283 \\
Tamil   &ta           &6224         &   447     &        832 \\
Thai    &th          &98064        &   2989      &      3713 \\
Turkish &tr         &182470       &    4045       &     5029 \\
Ukrainian    &uk         &108495       &    3060        &    3751 \\
Urdu    &ur          & 5977       &     508         &   1006 \\
Vietnamese  &vi         &171995       &    4645          &  4391 \\
Chinese-China &zh-cn      &199855       &    4558           & 5251 \\
Chinese &zh          & 5534       &     547            & 494 \\
Chinese-Taiwan &zh-tw      &202646      &     4583           & 5377 \\ \bottomrule

    \end{tabular}
    \caption{Languages and the parallel number of segments paired with English from the the TED Talks data~\cite{qi2018andPre-trainedEmbd}. The four languages used as an unseen ($U$) source are highlighted.}
    \label{tab:ted-talk-languages}
\end{table*}



\end{document}